\documentclass[11pt]{article}

\usepackage[final]{acl}
\usepackage{amsmath}  
\usepackage{times}
\usepackage{latexsym}
\usepackage{xspace}
\usepackage[T1]{fontenc}

\usepackage{microtype}

\usepackage{inconsolata}

\usepackage{multirow}
\usepackage{multicol}
\usepackage{booktabs}
\usepackage{subcaption}
\usepackage{hhline}
\usepackage{wasysym}
\usepackage{pifont}
\usepackage{algpseudocode}

\usepackage{enumitem}
\usepackage{adjustbox}
\usepackage{graphicx}
\usepackage{balance}
\usepackage{caption} 
\usepackage[most]{tcolorbox}
\usepackage{colortbl}
\usepackage{xcolor}    
\usepackage{amsthm}
\usepackage{anyfontsize}
\usepackage{bbm}
\usepackage{placeins}
\usepackage[normalem]{ulem}  
\usepackage{stfloats}

\newcommand{\paratitle}[1]{\vspace{0.8ex}\noindent \textbf{#1}}
\newcommand{\model}{HGP\xspace}
\renewcommand{\cite}{\citep}

\definecolor{RedOrange}{RGB}{255,69,0}

\definecolor{myup}{HTML}{A52A2A}

\newcommand{\figref}[1]{Figure~\ref{#1}}
\newcommand{\tabref}[1]{Table~\ref{#1}}

\title{HGP: An on-device personalized agent memory via hybrid graph storage}

\author{
  \textbf{Ran Zhou\textsuperscript{1}},
  \textbf{Xueming Han\textsuperscript{2}},
  \textbf{Jiaheng Liu\textsuperscript{1}},
  \textbf{Yuyao Zhang\textsuperscript{2}},
\\
  \textbf{Fanyu Meng\textsuperscript{2}},
  \textbf{Junlan Feng\textsuperscript{2}},
  \textbf{Yuxiang Ren\textsuperscript{1}\textsuperscript{\textdagger}}
\\
\\
  \textsuperscript{1}School of Intelligence Science and Technology, Nanjing University \\
  \textsuperscript{2}Jiutian Research \\
  \small{
    \textbf{Correspondence:} \texttt{renyuxiang@nju.edu.cn}
  }
}
\begin{document}
\maketitle
\footnotetext{Corresponding author: Yuxiang Ren.}
\begin{abstract}
LLM-based agents face challenges in personalized interactive tasks due to heterogeneous, multi-typed, and implicitly constrained long-term traces. Existing memory mechanisms struggle with accurate routing and retrieval, especially on-device where personalization is critical. Most methods use single-vector representations, blurring type distinctions and relational structure.
We propose \textbf{\model}, a hybrid graph memory framework. \model employs a lightweight self-enhancement classifier for personalized memory routing and constructs episodic, semantic, and procedural memories as graphs. It also extracts working memory as a state trajectory to capture current state and implicit constraints, ensuring reliable decision-making. The classifier reduces large-model calls, enabling on-device deployment, while graph storage enables accurate retrieval and incremental user profile refinement.
Experiments on two benchmarks show that on PAL-Set solution selection, \model{} achieves an S-score of 35.58, nearly 7 points above the strongest baseline. Code and data are at \url{https://github.com/Ouan6/HGP-.git}.
\end{abstract}

\section{Introduction}

LLM(Large Language Models)-based agents are increasingly deployed in personalized interactive tasks, such as long-term conversational companionship and constraint-aware task planning~\cite{yao2023reactsynergizingreasoningacting,park2023generativeagentsinteractivesimulacra}.
These applications require the agent to access user-related information across turns and generate responses that align with a specific user’s preferences and constraints~\cite{zhong2023memorybankenhancinglargelanguage,packer2024memgptllmsoperatingsystems}.
To compensate for limited context windows and unreliable long-context recall, external memory modules have been widely adopted~\cite{lewis2021retrievalaugmentedgenerationknowledgeintensivenlp,zhao2024rag,fan2024surveyragmeetingllms}.
Most agent memory systems adopt a unified memory pool, compress each memory item into a single-vector representation, retrieve by semantic similarity, and then concatenate retrieved snippets to support generation~\cite{lewis2021retrievalaugmentedgenerationknowledgeintensivenlp,zhong2023memorybankenhancinglargelanguage}.
However, such flat memory designs often blur distinctions among memory types and relational structure, which are critical for personalized retrieval and decision making.
Robust personalization demands continuously summarizing interaction experience, distilling user preferences and behavioral habits, and reliably activating such signals in subsequent decisions. These capabilities are essential for maintaining consistent user-specific behavior over long interactions
~\cite{park2023generativeagentsinteractivesimulacra,liu2023lostmiddlelanguagemodels}.
Moreover, a critical but underexplored bottleneck is personalized memory routing: dynamically assigning incoming interaction traces to the appropriate memory type.
In practice, histories contain heterogeneous signals (events, profiles, strategies, and actionable states)~\cite{atkinson1968human,tulving1972episodic,baddeley1974working,squire1992declarative}, and inaccurate routing stores information in the wrong memory type, causing cascading retrieval errors~\cite{xu2025amemagenticmemoryllm}.

\paratitle{Personalized memory framework is needed.}
We argue that achieving personalization requires a memory system with two key properties:
(a) enabling \emph{personalized memory routing} with low supervision cost; and
(b) explicitly modeling heterogeneous memory types and their relations.
First, personalized routing is important because memory distributions vary across users, making it difficult to rely solely on a fixed general-purpose classifier. Given the cost of repeatedly invoking LLMs during online interaction, especially in on-device contexts, a lightweight routing mechanism that amortizes supervisory cost over time becomes essential. In our setting, on-device operation refers to latency-critical online routing and retrieval on a local device, while graph construction and classifier updates are performed asynchronously. Second, an explicit modeling of memory heterogeneity is required: heterogeneous memory types must not be collapsed into a uniform vector space, but should instead retain their distinct attributes and interrelations through structured representations to facilitate retrieval.

\paratitle{Presented work.}
To address these challenges, we propose \model, a \textbf{H}ybrid \textbf{G}raph \textbf{P}ersonalized memory framework. Our main contributions are summarized as follows:
\FloatBarrier
\begin{itemize}[leftmargin=*]
    \item \textbf{Framework design.}
    We develop a framework that moves beyond single-vector compression by organizing episodic, semantic, and procedural memories with type-specific graph structures, including \emph{event graphs, knowledge graphs, and rule trees}. This is augmented by dynamic \emph{working memory-state trajectory graphs} that capture and preserve implicit contextual constraints.
    \item \textbf{Self-Enhancement Classifier.}
    We propose a confidence-based LLM-assisted annotation policy for personalized memory routing. A \emph{lightweight classifier} progressively takes over routing decisions, adapts to user-specific memory distributions, and substantially reduces supervision cost by recycling low-confidence LLM annotations as training data.
    \item \textbf{Empirical validation.}
     We evaluate \model on LoCoMo~\cite{maharana2024evaluatinglongtermconversationalmemory} and PAL-Set~\cite{huang2025mempalmemorybasedpersonalizeddialogue}, demonstrating consistent improvements in personalization quality and decision reliability over strong baselines. We also provide a thorough analysis of how working memory improves robustness under implicit constraints.
\end{itemize}

\section{Related Work}
\subsection{Long-context Modeling and Retrieval-Augmented Memory.}
One research direction improves model capacity by extending context windows and developing memory-optimized attention mechanisms, thereby enabling training and inference over significantly longer sequences.
Representative approaches include sparse-attention models such as Longformer~\cite{beltagy2020longformerlongdocumenttransformer} and BigBird~\cite{zaheer2021bigbirdtransformerslonger}, as well as I/O-efficient exact attention methods like FlashAttention~\cite{dao2022flashattentionfastmemoryefficientexact}.
Key innovations in positional encoding and length extrapolation, such as RoPE, ALiBi, and their derivatives~\cite{su2023roformerenhancedtransformerrotary,press2022trainshorttestlong,ding2024longropeextendingllmcontext}, have driven increases in the effective context window of LLM.
Representing a complementary strategy to long-context modeling, retrieval augmentation enhances LLMs by coupling them with an external knowledge store, as exemplified by RAG~\cite{lewis2021retrievalaugmentedgenerationknowledgeintensivenlp}, REALM~\cite{guu2020realmretrievalaugmentedlanguagemodel}, and RETRO~\cite{borgeaud2022improvinglanguagemodelsretrieving}.
However, increased context capacity does not guarantee effective information utilization. This limitation is evidenced by findings such as “Lost in the Middle”~\cite{liu2023lostmiddlelanguagemodels}, which demonstrate that models often access salient information unevenly within long contexts.
More critically, even with unbounded context length, both long-context modeling and retrieval augmentation lack the requisite mechanisms for continuously capturing and structuring user-specific interaction patterns, which are essential for genuine personalization.

\subsection{Graph-structured memory.}
Recent memory frameworks are increasingly adopting graph-based representations, driven by the promise of explicit relational reasoning~\cite{sun2025large,li2025advancing}. Moving beyond flat similarity search over text chunks, graph-structured memory enables more structured storage and retrieval through explicit modeling of relations among memory units.
GraphRAG-style systems~\cite{han2025retrievalaugmentedgenerationgraphsgraphrag,zhang2024lograg,zhang2024leveraging} index knowledge as graphs and retrieve query-relevant subgraphs to support multi-hop evidence aggregation.
Representative approaches include graph-traversal-based retrieval, such as HippoRAG~\cite{gutiérrez2025hipporagneurobiologicallyinspiredlongterm}, and hierarchical indexing with structure-aware retrieval, exemplified by LeanRAG~\cite{zhang2025leanragknowledgegraphbasedgenerationsemantic} and HiRAG~\cite{jiao2025hiraghierarchicalthoughtinstructiontuningretrievalaugmented}.
Graph-structured memory has also been explored in agentic settings, including G-Memory~\cite{zhang2025gmemorytracinghierarchicalmemory}, G-Refer~\cite{li2025grefergraphretrievalaugmentedlarge}, and the concurrent PlugMem~\cite{yang2026plugmemtaskagnosticpluginmemory}, which organizes agent experiences into structured memory graphs for efficient retrieval.
These works are closely related to HGP in moving beyond raw trajectory retrieval and flat memory storage.
However, most of them mainly focus on structure-aware retrieval or task-agnostic agent memory, while HGP focuses on personalized agent memory where heterogeneous user traces must be routed into distinct memory types and stored with type-specific graph structures for later personalized retrieval and decision making.

\subsection{Agent memory system.}
Memory is a core capability of LLM-based agents. 
Early systems mainly rely on passive retrieve-and-inject mechanisms such as retrieval-augmented generation (RAG)~\cite{lewis2021retrievalaugmentedgenerationknowledgeintensivenlp}, while agent frameworks like ReAct~\cite{yao2023reactsynergizingreasoningacting} and Toolformer~\cite{schick2023toolformerlanguagemodelsteach} treat memory as an implicit or short-term component.
To support personalization and long-term consistency, recent work increasingly incorporates cross-session memory. MemoryBank~\cite{zhong2023memorybankenhancinglargelanguage} continuously updates persistent user profiles; MemGPT~\cite{packer2024memgptllmsoperatingsystems} introduces a memory hierarchy between the context window and external storage; Generative Agents~\cite{park2023generativeagentsinteractivesimulacra} and MemoryOS~\cite{kang2025memoryosaiagent} further explore structured memory storage and lifecycle management.
Typed memory has also been discussed in cognitive agent architectures. For example, CoALA~\cite{sumers2024cognitivearchitectureslanguageagents} formalizes language agents with multiple memory components, including working, episodic, semantic, and procedural memories.
Production-oriented systems like Mem0~\cite{chhikara2025mem0buildingproductionreadyai} and A-MEM~\cite{xu2025amemagenticmemoryllm} target scalable memory pipelines, while recent work ScrapMem~\cite{chang2026scrapmembioinspiredframeworkondevice} further considers storage-efficient personalized memory on edge devices.
These works motivate persistent and cognitively inspired agent memory, whereas HGP studies how user traces can be routed into heterogeneous memory types and stored with type-specific graph structures for personalization.
Our self-enhanced router is related to cost-aware model routing and deferral methods, such as FrugalGPT~\cite{chen2023frugalgptuselargelanguage} and RouteLLM~\cite{ong2025routellmlearningroutellms}.
Unlike these methods, which route inputs across models, HGP routes memory writes into different memory stores; only low-confidence traces invoke LLM annotation, which is then cached for incremental classifier updates.

\begin{figure*}[t]
    \centering
    \includegraphics[width=\textwidth]{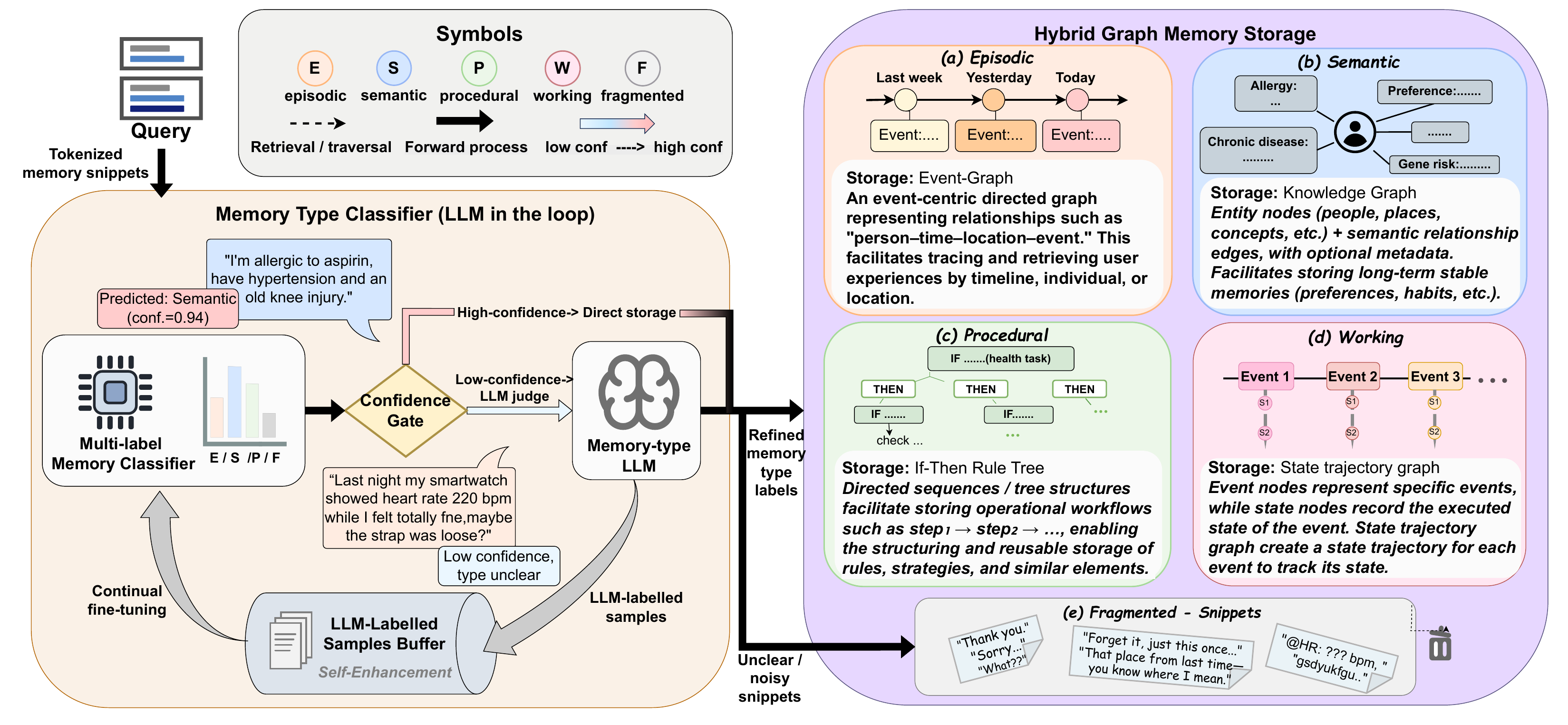}
    \vspace{-4mm}
    \caption{Overview of our \model framework with two components: (1) a self-enhancement classifier that routes inputs via a classifier/LLM and incrementally updates the classifier using high-confidence samples, and (2) hybrid graph storage that preserves structural properties of heterogeneous memory types. }
    \label{fig:hgp_graph}
    \vspace{-4mm}
\end{figure*}

\section{Methodology}
In this work, we propose \model, a hybrid graph memory framework for user personalization. As illustrated in Figure~\ref{fig:hgp_graph}, \model comprises two core components: (1) a \emph{self-enhancement classifier}; (2) a \emph{hybrid graph-based storage}.

\subsection{Personalized Memory Routing}
\label{sec:routing}
\label{sec:self_enhancement}
In cross-user scenarios, individual differences in memory-type signals cause substantial distribution shifts, making a general classifier prone to misrouting.
To address this, \model employs a confidence-based, LLM-assisted annotation mechanism that continuously refines the router and amortizes supervision cost.
For an input segment \(x\), the router first predicts a label \(\hat{c}_{\theta}(x)\) with confidence score \(s(x)\).
The final routing decision is governed by an iteration-adaptive threshold \(\tau_t\):
\begin{equation}
{c}(x) = 
\begin{cases}
\hat{c}_{\mathrm{LLM}}(x), & s(x) < \tau_t, \\
\hat{c}_{\theta}(x), & s(x) \ge \tau_t,
\end{cases}
\end{equation}
where \(\hat{c}_{\mathrm{LLM}}(x)\) denotes the label from the LLM-based router, and \(\tau_t\) is the confidence threshold at iteration \(t\).
If \(s(x) \ge \tau_t\), the predicted label \(\hat{c}_{\theta}(x)\) is retained for downstream memory construction.  
Otherwise, the segment is deferred to the LLM for annotation, and the resulting labeled pair \((x,\hat{c}_{\mathrm{LLM}}(x))\) is cached into an incremental training pool.
Periodically, the classifier is updated on the accumulated pool. 
We initialize the router conservatively (LLM-dominant) and gradually shift the workload to the classifier via exponential annealing:
\begin{equation}
\tau_t = \max\left(\tau_{\min}, \tau_0 \cdot e^{-\beta t} \right).
\end{equation}
where a high initial threshold \(\tau_0\) encourages deferral, a decay rate \(\beta\) enables a smooth transition, and a lower bound \(\tau_{\min}\) prevents overly aggressive self-labeling.
As the data pool grows with user-specific samples, the classifier's decision boundary progressively adapts to the current domain and user patterns, effectively mitigating cross-user distribution shifts.
This continuous adaptation sustains long-term personalization, which is especially critical as user behaviors evolve and memory distributions drift over time.
\renewcommand{\floatpagefraction}{.9}
\subsection{Hybrid Graph Storage}
\label{sec:hybrid_graph_store}
To preserve heterogeneous memory types and their relations, \model adopts a hybrid graph storage design. We detail each memory structure below.

\subsubsection{Episodic Memory -- Event Graph}
Episodic memory stores situated interaction events experienced, preserving traceable trajectories, temporal order, and contextual dependencies.
In \model, episodic memory is represented as an \emph{Event Graph}. Formally, the event graph is modeled as
\begin{equation}
\mathcal{G}_{\mathrm{epi}} = (V_e \cup V_{\mathrm{ent}}, E_{\mathrm{epi}}),
\end{equation}
where $V_e$ is the set of atomic episode nodes, $V_{\mathrm{ent}}$ denotes contextual entity nodes, and $E_{\mathrm{epi}}$ represents typed edges encoding event–event and event–entity relations.

In the event graph, episode nodes serve as structural anchors, connected to other episodes or contextual entities via semantically typed edges. These edges capture structured dependencies between events, such as temporal order, contextual association, and other semantic relations. Each edge is defined as
\begin{equation}
e_{ij} = (v_i, v_j, r), \quad r \in \{\textit{temporal},  \textit{contextual}, \dots\},
\end{equation}
where $r$ specifies the semantic role of the relation between two nodes. Unlike isolated record storage, the event graph links events through shared entities, preserving structured relationships and contextual continuity.

As a core component of the hybrid graph storage, the event graph further supports structure-aware retrieval and long-term evolution. As event graphs accumulate, recurring relational patterns and substructures can be abstracted as:
\begin{equation}
\mathcal{K} = \mathcal{A}_{\mathrm{abs}}\big({\mathcal{G}_{\mathrm{epi}}^{(1)}, \dots, \mathcal{G}_{\mathrm{epi}}^{(n)}}),
\end{equation}
where $\mathcal{A}{\mathrm{abs}}(\cdot)$ denotes a structure abstraction operator instantiated via LLM. Through this process, episodic memory functions not only as a repository of past experiences but also as a structural foundation for deriving higher-level semantic and procedural knowledge.

\subsubsection{Semantic Memory -- Knowledge Graph}
Semantic memory captures long-term, decontextualized knowledge such as facts, relations, and user preferences, supporting consistent personalized behavior. In \model, it is explicitly represented as a \emph{knowledge graph}:
\begin{equation}
\mathcal{G}_{\mathrm{sem}} = (V_{\mathrm{sem}}, E_{\mathrm{sem}}),
\end{equation}
where $V_{\mathrm{sem}}$ denotes semantic nodes encoding user-centric long-term knowledge, and $E_{\mathrm{sem}}$ represents typed semantic relations among them.

Each semantic relation is represented as a triple:
\begin{equation}
(u, r, v), \quad 
u \in V_{\mathrm{user}},\;
v \in V_{\mathrm{ent}},\;
r \in \mathcal{R}_{\mathrm{sem}},
\end{equation}
where $r$ is a predicate drawn from a predefined semantic relations (e.g., \textsc{HAS\_SKILL}, \textsc{PREFERS}, \textsc{WORKS\_AS}). 

Unlike episodic memory, which captures temporally grounded experience fragments, semantic memory focuses on regularities that generalize across interaction contexts. It can be obtained either directly via the self-enhancement classifier or through semantic abstraction over accumulated episodic event graphs:
\begin{equation}
\mathcal{G}_{\mathrm{sem}} = 
\mathcal{A}_{\mathrm{sem}}
\big(
\mathcal{G}_{\mathrm{epi}}^{(1)}, \dots, \mathcal{G}_{\mathrm{epi}}^{(n)}
\big).
\end{equation}
The detailed design and prompt templates of $\mathcal{A}_{\mathrm{sem}}$ are provided in Appendix~\ref{appendix:prompt_sem}.

\subsubsection{Procedural Memory -- Rule Tree Graph}
Procedural memory stores reusable, experience-driven behavioral strategies and decision flows, enhancing long-term planning capability. In \model, procedural memory is structured as a \emph{Rule Tree}, where nodes represent conditions, actions, or intermediate goals, and edges encode their dependencies. Formally, it is defined as
\begin{equation}
\mathcal{G}_{\mathrm{proc}} = (V_c \cup V_a \cup V_g, E_{\mathrm{proc}}),
\end{equation}
where $V_c$, $V_a$, $V_g$ denote condition, action, and sub-goal nodes, respectively, and \(E_{\mathrm{proc}}\) represents the set of edges encoding control or dependency relations among these nodes.

Procedural memory is sourced directly from the self-enhancement classifier or extracted from episodic event graphs. In the latter case, it is constructed via
\begin{equation}
\mathcal{G}_{\mathrm{proc}} = \mathcal{A}_{\mathrm{proc}}\big(\mathcal{G}_{\mathrm{epi}}^{(1)}, \dots, \mathcal{G}_{\mathrm{epi}}^{(n)}),
\end{equation}
where $\mathcal{A}_{\mathrm{proc}}(\cdot)$ is a procedural-abstraction operator instantiated with an LLM. The detailed design and prompt template of $\mathcal{A}_{\mathrm{proc}}$ are provided in Appendix~\ref{appendix:prompt_proc}.

\subsubsection{Working Memory -- State Trajectory Graph}
Working memory maintains the states of ongoing interactions or tasks, supplying state information for reasoning and decision-making. In \model, it is implemented as a \emph{State Trajectory Graph}, where nodes represent current system or user states and edges encode state transitions. Formally, working memory is defined as
\begin{equation}
\mathcal{G}_{\mathrm{wm}} = (V_{\mathrm{state}}, E_{\mathrm{state}}),
\end{equation}
where $V_{\mathrm{state}}$ represents the current states and $E_{\mathrm{state}}$ denotes transitions between states.

Unlike episodic and semantic memory, which retain long-term information, working memory maintains short-lived, dynamic context extracted from recent episodic memory and continuously updated during interaction. Its update process can be expressed as
\begin{equation}
\mathcal{G}_{\mathrm{wm}}^{(t+1)} = \mathcal{U}\big(\mathcal{G}_{\mathrm{wm}}^{(t)}, \mathcal{G}_{\mathrm{epi}}^{(t)}\big),
\end{equation}
where $\mathcal{U}(\cdot)$ denotes a state update operator. Graph-based modeling of working memory provides an explicit representation of the current state and its evolution, enabling timely reasoning, strategy selection, and action triggering.

\subsubsection{Query Retrieval Workflow}
Upon receiving a user query \( q \), \model\ executes a structured retrieval protocol over its hybrid graph memory \( \mathcal{G} = \{ \mathcal{G}_{\mathrm{epi}}, \mathcal{G}_{\mathrm{sem}}, \mathcal{G}_{\mathrm{proc}}, \mathcal{G}_{\mathrm{wm}} \} \). A detailed example of this retrieval workflow is provided in Appendix~\ref{appendix:retrieval_example}.
\noindent
For the query \( q \), \model\ issues parallel queries to each subgraph, each designed to retrieve a particular type of structured evidence:
\begin{itemize}[leftmargin=*,noitemsep]
    \item \textbf{Episodic graph \( \mathcal{G}_{\mathrm{epi}} \)}: Retrieves a set of temporally or contextually related past interaction nodes. For instance, in a health management scenario, a query about “evening exercise planning” may retrieve nodes representing \textit{``night run with elevated heart rate''} and \textit{``previous over-exertion alert''}, connected via shared activity and time entities.
    \item \textbf{Semantic graph \( \mathcal{G}_{\mathrm{sem}} \)}: Retrieves a subgraph of factual triples about the user that are semantically relevant to \( q \). Examples include \textit{(User, allergy, Ibuprofen)} and \textit{(User, condition, Hypertension)}, providing stable profile constraints.
    \item \textbf{Procedural graph \( \mathcal{G}_{\mathrm{proc}} \)}: Retrieves applicable rule nodes that match the current context. For example, the rule \textit{``If NSAID combines with aspirin → check bleeding risk''} is retrieved when the query involves taking painkillers after aspirin intake.
    \item \textbf{Working memory graph \( \mathcal{G}_{\mathrm{wm}} \)}: Retrieves the current state trajectory, capturing immediate session constraints such as \textit{``aspirin already taken this morning''} and \textit{``pending dinner recommendation task''}.
\end{itemize}

\noindent
The retrieved subgraphs are merged into a unified context and passed, together with the original query \( q \), to the LLM.
The model then generates a response grounded in both long-term memory and the user’s current state and implicit constraints, ensuring coherence, personalization, and safety.

\begin{table*}[tb]
\centering
\small
\setlength{\tabcolsep}{3.6pt}
\renewcommand{\arraystretch}{1.15}
\caption{Unified evaluation results on LoCoMo (\%). 
Evaluation metrics are reported for each category and overall, including F1 score (F1), BLEU-1 (B1), Exact Match score (EM) and LLM-as-a-Judge score (J), with higher values indicating
better performance. Best results are in \textbf{bold}, and second-best results are \uline{underlined}.}
\label{tab:locomo_unified_bigtable_percent}

\begin{adjustbox}{width=\textwidth}
\begin{tabular}{l||cccc||cccc||cccc||cccc||cccc}
\toprule
\multirow{2}{*}{\textbf{Method}} 
& \multicolumn{4}{c||}{\textbf{Single-hop}} 
& \multicolumn{4}{c||}{\textbf{Multi-hop}} 
& \multicolumn{4}{c||}{\textbf{Temporal}} 
& \multicolumn{4}{c||}{\textbf{Open-domain}} 
& \multicolumn{4}{c}{\textbf{Overall}} \\
\cmidrule(lr){2-5} \cmidrule(lr){6-9} \cmidrule(lr){10-13} \cmidrule(lr){14-17} \cmidrule(lr){18-21}
& \textbf{J} & \textbf{F1} & \textbf{B1} & \textbf{EM}
& \textbf{J} & \textbf{F1} & \textbf{B1} & \textbf{EM}
& \textbf{J} & \textbf{F1} & \textbf{B1} & \textbf{EM}
& \textbf{J} & \textbf{F1} & \textbf{B1} & \textbf{EM}
& \textbf{J} & \textbf{F1} & \textbf{B1} & \textbf{EM} \\
\midrule

TurnRAG
& 47.52 & 34.31 & 28.43 & 6.03
& 38.54 & 20.33 & 16.60 & 6.25
& 60.44 & 51.71 & \textbf{47.86} & \textbf{19.94}
& 65.28 & 48.34 & 43.76 & 24.97
& 59.35 & 44.73 & 40.11 & 19.29 \\
\midrule

LangMem
& 53.55 & 32.78 & 26.35 & 3.90
& 41.67 & 22.97 & 19.99 & 10.42
& 17.76 & 10.23 & 8.67 & 3.74
& 54.22 & 34.30 & 30.64 & 14.10
& 45.65 & 24.74 & 21.49 & 8.46 \\
MemoryBank
& 47.87 & 19.26 & 24.29 & 3.55
& 37.50 & 10.89 & 15.00 & 7.29
& 26.17 & 6.63 & 7.51 & 4.05
& \textbf{72.65} & 40.97 & 44.37 & 26.27
& 56.12 & 39.83 & 35.03 & 16.22 \\
A-Mem
& 51.42 & 20.43 & 25.29 & 4.96
& 44.79 & 13.60 & 17.00 & 7.29
& 59.19 & 35.21 & 38.63 & 19.31
& 67.59 & 38.42 & 41.09 & 24.82
& 61.41 & 32.87 & 36.15 & 18.90 \\
HippoRAG2
& 52.84 & 21.16 & 27.69 & 4.61
& 40.62 & 13.25 & 17.08 & 8.33
& \textbf{60.75} & 38.00 & 41.75 & \textbf{19.94}
& 68.31 & 39.75 & 42.84 & 26.39
& 62.13 & 34.29 & 38.20 & 19.88 \\
Mem0
& 52.13 & \textbf{35.18} & 26.15 & \textbf{5.67}
& 38.54 & \textbf{28.95} & 22.85 & \textbf{13.54}
& 59.81 & \textbf{52.71} & 39.70 & 19.31
& 66.39 & 47.98 & 40.81 & 26.14
& 60.63 & \textbf{45.42} & 36.75 & 20.14 \\
\midrule

\textbf{Ours}
& \textbf{56.03} & \uline{34.44} & \textbf{30.39} & \textbf{5.67}
& \textbf{46.88} & \uline{27.69} & \textbf{23.70} & \uline{12.50}
& \textbf{60.75} & \uline{50.31} & \uline{46.24} & \uline{19.63}
& \uline{71.45} & \textbf{51.33} & \textbf{46.59} & \textbf{26.87}
& \textbf{65.08} & \uline{45.07} & \textbf{42.09} & \textbf{20.54} \\
\bottomrule
\end{tabular}
\end{adjustbox}
\end{table*}

\section{Experiments}
\subsection{Experimental Setup}
\paratitle{Datasets.}
We evaluate on two long-term memory benchmarks:
LoCoMo~\cite{maharana2024evaluatinglongtermconversationalmemory} provides 10 multi-session conversations with $\sim$200 QA pairs each, covering single-hop, multi-hop, temporal, and open-domain questions. Following Mem0~\cite{chhikara2025mem0buildingproductionreadyai}, we exclude the adversarial category due to missing ground-truth answers.
PAL-Set~\cite{huang2025mempalmemorybasedpersonalizeddialogue} evaluates memory-aware personalization via user logs and interaction histories. We focus on the \textit{Solution Proposal} tasks: (1)QA: memory-grounded answer generation; (2)Selection: selecting the best response from candidates based on retrieved memories.

\paratitle{Metrics.}
On LoCoMo, we report F1, Exact Match (EM), BLEU-1 (B1), and supplement with Mem0-style LLM-as-a-judge(J) to overcome the limits of lexical overlap in long-context reasoning.
On PAL-Set, for the \textit{QA} subtask, we use BLEU-1/2/3/4 as automatic metrics and report an LLM-as-a-judge score as a complementary assessment of answer correctness and usefulness.
For comparability, the same judging protocol is applied to both memory-based and non-memory systems.
For \textit{Selection}, we use the benchmark-defined S-score, which measures how well the selected solution matches the user's needs.
The LLM-as-a-judge score is treated as supplementary evidence rather than the sole evaluation criterion; our conclusions are based jointly on automatic QA metrics, the PAL-Set S-score, ablation results, and long-horizon memory evaluation.
Full QA judge prompts are provided in Appendix~\ref{appendix:prompt}.

\paratitle{Baselines.}
We compare against two paradigms:
\begin{enumerate}[label=(\arabic*), leftmargin=*, topsep=2pt]
    \item \textbf{Non-memory methods.} 
    On LoCoMo, we use a standard RAG pipeline. 
    On PAL-Set, following the official benchmark, we include four retrieval-based variants with different context granularities: 
    \textit{Vanilla (w/o log)} (no history), 
    \textit{Vanilla (with log)} (logs concatenated into the prompt), 
    \textit{Turn-level RAG}, and 
    \textit{Session-level RAG}.
    \item \textbf{Memory-based frameworks.} 
    We compare with representative memory-augmented systems: 
    Mem0~\cite{chhikara2025mem0buildingproductionreadyai}, 
    MemoryBank~\cite{zhong2023memorybankenhancinglargelanguage}, 
    A-Mem~\cite{xu2025amemagenticmemoryllm}, 
    HippoRAG2~\cite{gutiérrez2025hipporagneurobiologicallyinspiredlongterm}, 
    and additionally LangMem on LoCoMo.
\end{enumerate}

\paratitle{Implementation Details.}
In \model, memory routing is performed by a lightweight BERT-based classifier (\texttt{roberta\_L-4\_H-512}). The confidence threshold is initialized at $\tau_0 = 0.90$, annealed with rate $\beta = 0.05$, and lower-bounded by $\tau_{\min} = 0.50$. Since working-memory annotations are not available in LoCoMo or PAL-Set, our experiments focus on episodic, semantic, and procedural memories, stored as Neo4j graph databases.
Memory extraction uses \texttt{deepseek-chat}; response generation and LLM-as-a-judge use \texttt{gpt-4o-mini} (temperature $0.0$). On LoCoMo, max output length is 100 tokens; dialogue history for TurnRAG is truncated to 4K tokens; retrieval-based methods use $K = 30$ items. All methods share the same answer/judge prompts adapted from Mem0 for fairness.
On PAL-Set, we follow the official setup with $K = 3$ and introduce an additional QA-specific judge prompt for semantic correctness beyond lexical overlap.
\begin{table*}[t]
\centering
\small
\setlength{\tabcolsep}{6pt}
\renewcommand{\arraystretch}{1.12}
\caption{Performance comparison across different methods on PAL-Set Solution QA tasks. B-1 to B-4 represent BLEU scores. $\Delta$ (vs. Vanilla w/o log) reports the absolute improvement in S-score compared to the Vanilla (w/o log) baseline. LLM-judge denotes the score evaluated by GPT-4o-mini.}
\label{tab:main_results}

\begin{tabular}{ll|cccc|c||cc}
\toprule
\textbf{Group} & \textbf{Method}
& \textbf{B-1} & \textbf{B-2} & \textbf{B-3} & \textbf{B-4}
& \textbf{QA Judge} & \textbf{S-score} & \textbf{$\Delta$ (vs. Vanilla w/o log)} \\
\midrule

\multirow{4}{*}{\textit{Non-memory}}
& Vanilla (w/o log) & 19.68 & 7.72 & 3.80 & 2.11 & 7.31 & 20.34 & +0.00 \\
& Vanilla (with log) & 18.96 & 7.51 & 3.82 & 2.22 & 7.32 & 25.00 & +4.66 \\
& Turn-level RAG & 19.48 & 7.72 & 3.93 & 2.23 & 7.41 & 25.06 & +4.72 \\
& Session-level RAG & 19.17 & 7.55 & 3.82 & 2.22 & 7.30 & 25.24 & +4.90 \\
\midrule
\addlinespace[0.25ex]
\midrule

\multirow{5}{*}{\textit{Memory-based}}
& MemoryBank & \textbf{20.83} & 7.36 & 4.23 & 2.56 & 7.13 & 28.85 & +8.51 \\
& Mem0 & 19.20 & 7.66 & 3.82 & 2.12 & 7.34 & 26.50 & +6.16 \\
& A-Mem & 19.33 & 8.12 & 3.96 & 2.17 & 7.04 & 27.27 & +6.93 \\
& HippoRAG2 & 18.84 & 8.06 & 4.30 & 2.34 & 7.31 & 29.22 & +8.88 \\
\cmidrule(lr){2-9}
& \textbf{HGP (Ours)} & 19.62 & \textbf{8.67} & \textbf{4.70} & \textbf{2.78} & \textbf{7.46} & \textbf{35.58} & \textbf{+15.24} \\
\bottomrule
\end{tabular}
\end{table*}

\subsection{Main Results}
\paratitle{LoCoMo performance.}
Experiments on the LoCoMo dataset demonstrate the strong memory recall capability of \model.
As shown in \tabref{tab:locomo_unified_bigtable_percent}, \model achieves the best overall performance in Judge (65.08), BLEU-1 (42.09), and EM (20.54), while maintaining competitive F1 (45.07)—outperforming Mem0 by +4.45 (Judge) and +5.34 (BLEU-1).
Gains are most pronounced on Open-domain tasks, where \model attains the highest F1 (51.33), BLEU-1 (46.59), and EM (26.87). This improvement supports the effectiveness of graph-based memory storage: structuring episodic events, semantic profiles, and procedural rules as interconnected graphs provides more reliable evidence for diverse, unconstrained questions.
Temporal questions remain comparatively challenging, suggesting opportunities for enhancing temporal ordering.

\paratitle{PAL-Set Solution Proposal tasks.}
Table~\ref{tab:main_results} reports results on Solution QA and Selection.
On QA, \model achieves the highest LLM-judge score (7.46) and the best BLEU-2/3/4 (8.67/4.70/2.78), indicating stronger multi-step coherence.
On Selection, non-memory baselines improve only marginally over Vanilla (w/o log) ($+4.66$ to $+4.90$), confirming that raw log retrieval is insufficient. \model attains an S-score of 35.58, surpassing Vanilla (w/o log) (20.34) by $+15.24$ and the top memory baseline HippoRAG2 (29.22) by $+6.36$.
These results show that graph-structured memory enables more precise personalized retrieval, better aligning responses with user-specific needs.
\begin{figure*}[t]
    \centering
    \includegraphics[width=\textwidth]{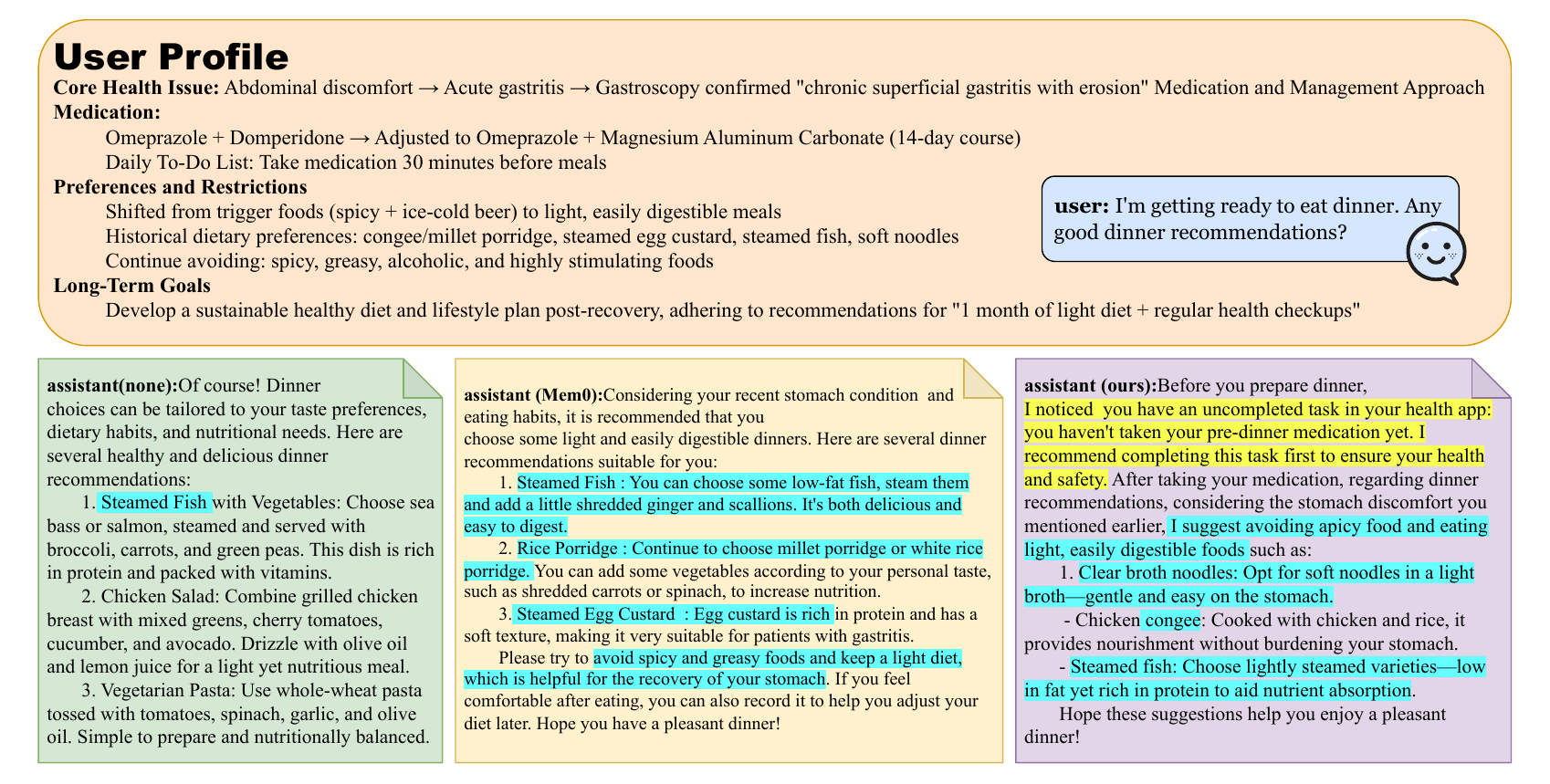}
    \vspace{-4mm}
    \caption{Working Memory Analysis: working-memory state tracking in a personal healthcare scenario. Yellow highlights denote pending to-do reminders, and blue highlights indicate preference-aligned personalized suggestions.
Compared with baselines, our method surfaces the working-memory state (pre-meal medication: pending) and reminds the user while providing personalized dinner recommendations.}
    \label{fig:case}
    \vspace{-4mm}
\end{figure*}

\subsection{Ablation Study}
\begin{table}[t]
\centering
\small
\setlength{\tabcolsep}{6pt}
\renewcommand{\arraystretch}{1.12}
\caption{Ablation study on memory components and representation designs. The table reports generation quality (BLEU and LLM-judge) and selection performance (S-score). Higher is better.}
\label{tab:ablation_memory}

\resizebox{\columnwidth}{!}{%
\begin{tabular}{l|ccccc|c}
\toprule
\textbf{Method} 
& \textbf{B-1} & \textbf{B-2} & \textbf{B-3} & \textbf{B-4} 
& \textbf{LLM-judge} & \textbf{S-score} \\
\midrule
Unified-vector
& 18.15 & 7.24 & 3.60 & 2.05 & 7.18 & 25.15 \\
Typed-vector
& 18.61 & 7.51 & 3.74 & 2.09 & 7.29 & 33.66 \\
\midrule
w/o \textit{E-Mem} 
& 19.53 & 7.96 & 3.96 & 2.23 & 7.09 & 28.83  \\
w/o \textit{S-Mem} 
& \textbf{19.69} & 8.30 & 4.38 & 2.57 & 7.23 & 25.71 \\
w/o \textit{P-Mem} 
& 19.59 & 8.63 & 4.66 & 2.76 & 7.31 & 33.10 \\
\midrule
\textbf{Full (Ours)} 
& \uline{19.62} & \textbf{8.67} & \textbf{4.70} & \textbf{2.78} 
& \textbf{7.46} & \textbf{35.58} \\
\bottomrule
\end{tabular}%
}
\end{table}
To quantify the contribution of each memory component, We ablate each memory type while keeping the rest of the pipeline unchanged (Table~\ref{tab:ablation_memory}).
\paratitle{Graph and type structure.}
Unified-vector collapses all memories into a single vector space, while Typed-vector preserves memory types but removes graph structure.
Typed-vector substantially improves over Unified-vector on \textit{Selection} (33.66 vs. 25.15), showing that memory typing is important for personalized decision making.
Full HGP further improves the S-score to 35.58, indicating that type-specific graph storage provides additional gains beyond typed vector retrieval.

\paratitle{Episodic Memory(E-Mem).}
Removing episodic memory causes the largest drop on \textit{Solution QA}: BLEU-1/2/3/4 fall from 19.66/8.67/4.70/2.78 to 19.53/7.96/3.96/2.23, and the LLM-judge score declines from 7.46 to 7.09. On \textit{Selection}, the S-score drops from 35.58 to 28.83 ($\Delta = -6.75$). These results confirm that episodic memory, as the foundational trace of user interactions, is essential for both precise answer generation and context-aware solution matching.

\paratitle{Semantic Memory(S-Mem).}
Removing semantic memory causes the largest drop on \textit{Selection} ($\Delta = -9.87$,S-score; Table~\ref{tab:ablation_memory}), confirming that stable user profiles (long-term preferences, attributes, health facts) are essential for personalized decision-making. On \textit{Solution QA}, the LLM-judge score also declines (7.46 to 7.23), indicating its contribution to coherent memory recall beyond pure retrieval.

\paratitle{Procedural Memory(P-Mem).}
Removing procedural memory causes a modest but non-negligible drop in \textit{Selection} ($\Delta = -2.48$), while the impact on \textit{Solution QA} is minor (judge score $7.46\rightarrow7.31$, BLEU-4 $2.78\rightarrow2.76$). This aligns with its role as high-level strategic guidance rather than direct factual evidence.

Overall, Table~\ref{tab:ablation_memory} highlights complementary roles: semantic graphs drive personalized selection, episodic graphs supply fine-grained event evidence for QA, and procedural graphs enhance strategy reuse. Integrating all components yields the best performance, validating our hybrid graph memory design.

\label{sec:special_case_study}
\subsection{Working Memory Analysis}
\figref{fig:case} illustrates the role of \textit{working memory} in \model using a healthcare scenario. Working memory explicitly extracts and maintains the current actionable state, enabling the agent to prioritize immediate actions during response generation.

\paratitle{Setup.}
We compare three systems: (i) a vanilla LLM without memory, (ii) an LLM augmented with Mem0, and (iii) \model with working memory enabled. The dialog history contains a pre-meal medication reminder. The user query asks for dinner recommendations without explicitly mentioning medication.

\paratitle{Observation.}
The vanilla LLM directly suggests dinner options. Mem0 generates preference-aligned recommendations but fails to remind the user about the medication. In contrast, \model\ retrieves the state ``pre-meal medication: unfinished'' from the \emph{working memory graph} and injects it into the prompt, enabling the LLM to proactively issue a medication reminder before dinner and demonstrating state awareness under implicit constraints. (Further details in Appendix~\ref{appendix:special_case}.)

\begin{figure}[t]
    \centering
    \resizebox{1.0\columnwidth}{!}{\includegraphics{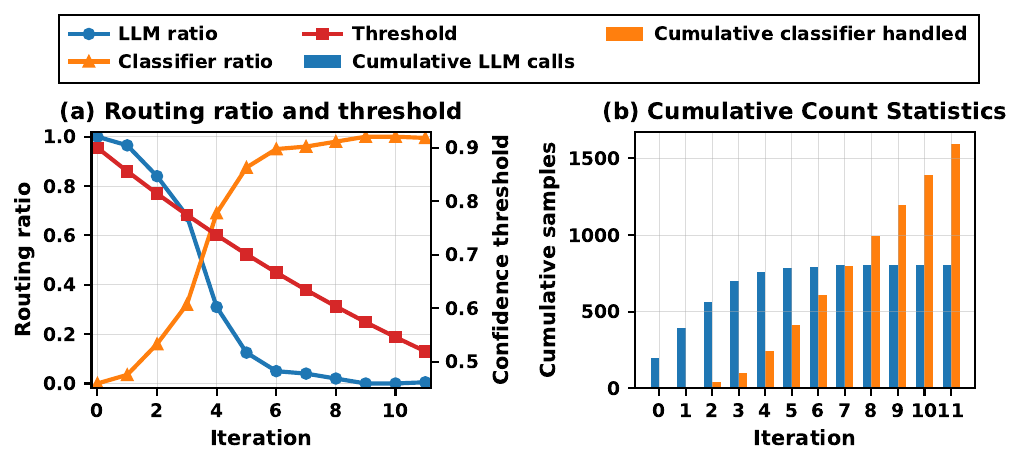}}
    \vspace{-6mm}
    \caption{Classifier routing statistics across iterations. (a) Routing mix and adaptive threshold. (b) Cumulative workload handled by the LLM and the classifier.}
    \label{fig:routing}
    \vspace{-4mm}
\end{figure}
Table~\ref{tab:router_results} compares our self-enhancement classifier with an offline classifier, a fixed LLM router, and a zero-shot LLM router. Our classifier achieves the highest macro-F1 (0.9339) and micro-F1 (0.9105) with an average latency of 4.98 ms, roughly 360× faster than the fixed LLM router. 
Fig.~\ref{fig:routing} shows the training dynamics: the confidence-based gating progressively shifts routing decisions from the LLM to the classifier, and incremental training every 200 samples drives the system from LLM-dominant to classifier-dominant (Fig.~\ref{fig:routing}a). 
Overall, LLM calls drop from 2400 to 807 (a 66.4\% reduction), confirming the workload transfer visible in the cumulative curves (Fig.~\ref{fig:routing}b). 
These results demonstrate that the classifier effectively captures LLM routing knowledge while substantially lowering inference costs for on-device deployment. 
Hyperparameter sensitivity and per-stage performance are detailed in Appendix~\ref{appendix:classifier}.
\begin{table}[t]
\centering
\small
\caption{Performance comparison of the self-enhancement classifier against other baselines. Best results are in \textbf{bold}, and second-best results are \uline{underlined}.}
\label{tab:router_results}

\resizebox{\columnwidth}{!}{%
\begin{tabular}{lccc}
\toprule
\textbf{Router} & \textbf{Micro-F1} & \textbf{Macro-F1} & \textbf{Avg Latency (ms)} \\
\midrule
Offline Classifier  & 0.8638 & 0.8790 & \textbf{4.06} \\
Fixed LLM Router    & 0.8953 & 0.9224 & 1803.69 \\
Zero-shot LLM Router& 0.8286 & 0.8436 & 1572.64 \\
\textbf{Ours}   & \textbf{0.9105} & \textbf{0.9339} & \uline{4.98} \\
\bottomrule
\end{tabular}%
}
\end{table}
\vspace{1.15mm}
\section{Conclusion}
In this paper, we address the key challenge of achieving robust and adaptive personalization in LLM-based agents under cross-user heterogeneity and distribution shift. To this end, we propose \model, a memory-augmented framework that integrates a self-enhancement routing mechanism with a hybrid graph-based memory storage, enabling cost-efficient and structured personalization.
Extensive experiments on LoCoMo and PAL-Set validate the effectiveness of \model in improving memory-recall accuracy and personalization capability, while routing statistics confirm its ability to amortize supervision cost over time.
Future work will explore finer-grained modeling of memory graphs, incorporate cross-memory consistency constraints, and extend the framework to a broader range of interactive tasks and tool-augmented agents.

\section{Limitations}
While \model demonstrates improvements in personalized memory management, several limitations remain.
First, although working memory shows promising qualitative results, existing benchmarks lack explicit working-memory annotations, preventing systematic quantitative evaluation. Constructing datasets with ground-truth state trajectories is an important direction for rigorous assessment.
Second, graph construction and maintenance introduce additional offline computation and storage costs, while our on-device claim primarily concerns latency-critical online routing and retrieval. A more complete end-to-end cost analysis remains for future work.
In addition, this work focuses mainly on graph storage, while finer-grained retrieval and pruning mechanisms require further study.
Finally, we use LLM-as-a-judge evaluation to complement lexical-overlap metrics. Although standardized prompts and temperature 0.0 improve evaluation consistency, LLM judges may still exhibit biases, and their reliability warrants continued attention.
\section*{Acknowledgments}
This work was  supported by the Nanjing University-China Mobile Joint Research Institute.


\bibliography{custom}

\appendix

\section{More Experiment Details}
\subsection{Prompts for PAL-Set Solution QA}
\label{appendix:prompt}
\begin{figure*}[t!]
    \centering
    \includegraphics[width=\textwidth]{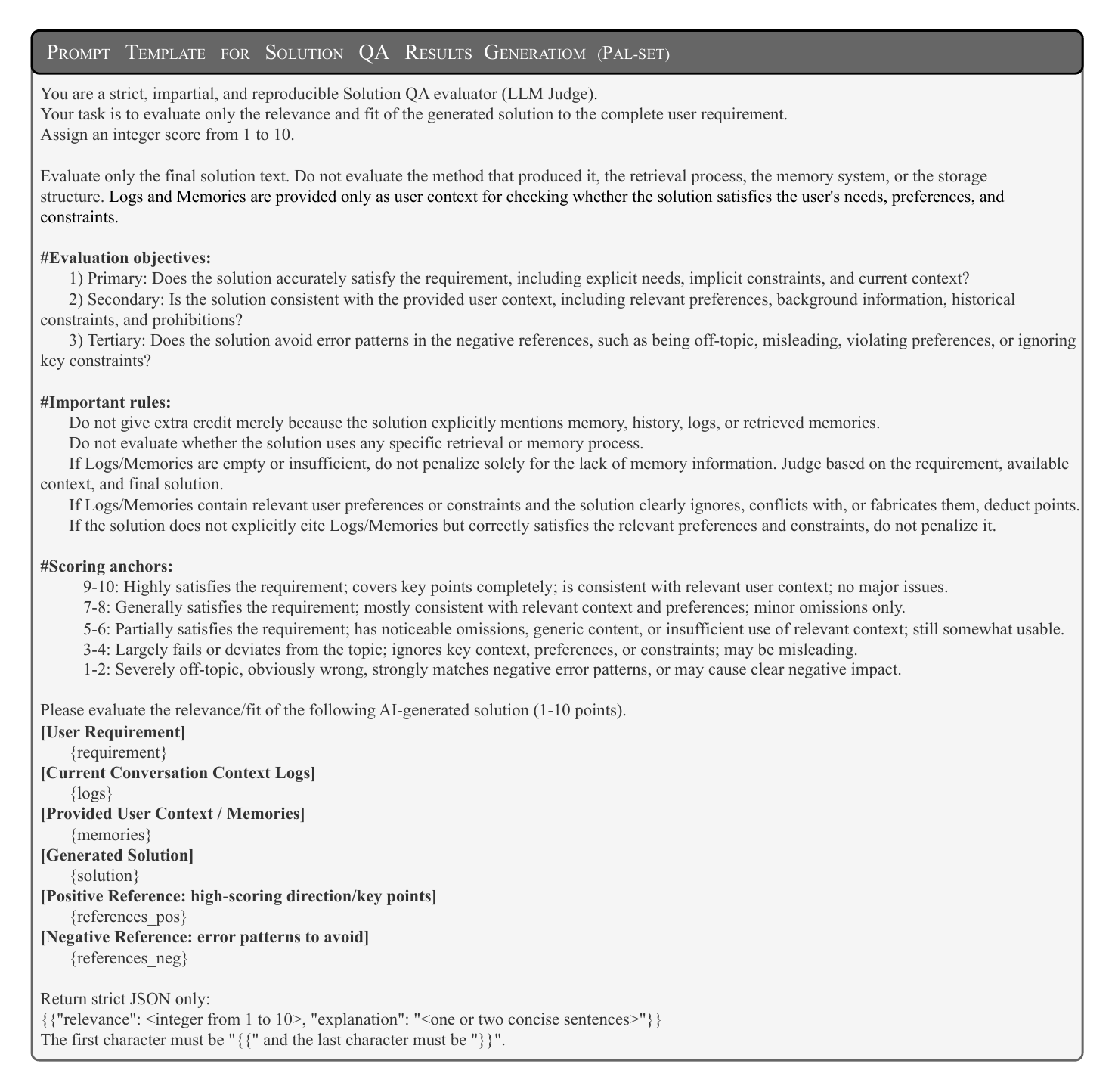}
    \vspace{0mm}
    \caption{LLM-as-a-Judge prompt used for PAL-Set Solution QA evaluation, enforcing requirement-fit scoring with implicit constraints, retrieved memories/logs, and pos/neg reference calibration.}
    \label{fig:prompt}
    \vspace{0mm}
\end{figure*}

 We provide the full LLM-as-a-Judge prompt template for the PAL-Set Solution QA task in Figure~\ref{fig:prompt}. This template is used to compute the LLM-judge scores reported in \tabref{tab:main_results}. During evaluation, we supply the judge LLM with the model-generated solution, the task question, and relevant conversation context (when available), along with the corresponding positive and negative reference solutions from the dataset.
The judge is instructed to score outputs across multiple dimensions: requirement satisfaction, semantic correctness, step completeness, and user-level personalization. Positive and negative references are provided to calibrate judgments and penalize common error patterns. For ease of aggregation, the judge is constrained to produce strictly formatted JSON outputs, with temperature fixed at 0.0.
\label{app:palset_judge_prompt}

\subsection{Supplementary Experiments on the Self-Enhancement Classifier}
\label{appendix:classifier}
\subsubsection{Hyperparameter Sensitivity Analysis}
\begin{table*}[t]
\centering
\normalsize
\setlength{\tabcolsep}{8pt}
\renewcommand{\arraystretch}{1.1}
\caption{Hyperparameter sensitivity analysis of the router with different values of $\tau_0$, $\beta$, and $\tau_{\min}$. The configuration selected for our experiments is highlighted in bold.}
\label{tab:sensitivity}
\begin{tabular}{cccccccc}
\toprule
Rank & $\tau_0$ & $\beta$ & $\tau_{\min}$ & \multicolumn{1}{c}{F1 (mean$\pm$std)} & \multicolumn{1}{c}{Latency (ms)} & Norm Latency & Score \\
\midrule
1  & 0.9  & 0.02 & 0.5 & $0.9490\pm0.0034$ & 4.23 & 1.000 &0.947 \\
2  & 0.9  & 0.02 & 0.3 & $0.9490\pm0.0034$ & 4.36 & 1.031 & 0.932 \\
3  & 0.9  & 0.02 & 0.7 & $0.9490\pm0.0034$ & 4.40 & 1.040 & 0.927 \\
4  & 0.8  & 0.1  & 0.7 & $0.9482\pm0.0023$ & 4.46 & 1.055 & 0.919 \\
\textbf{5}  & \textbf{0.9}  & \textbf{0.05} & \textbf{0.5} &\textbf{ $0.9474\pm0.0060$} & \textbf{4.44} & \textbf{1.050} & \textbf{0.919} \\
6  & 0.8  & 0.02 & 0.5 & $0.9482\pm0.0023$ & 4.72 & 1.118 & 0.888 \\
7  & 0.9  & 0.05 & 0.3 & $0.9474\pm0.0060$ & 4.83 & 1.143 & 0.873 \\
8  & 0.95 & 0.1  & 0.5 & $0.9526\pm0.0001$ & 4.96 & 1.174 & 0.866 \\
9  & 0.9  & 0.1  & 0.5 & $0.9474\pm0.0060$ & 5.06 & 1.198 & 0.846 \\
10 & 0.8  & 0.05 & 0.3 & $0.9482\pm0.0023$ & 5.10 & 1.207 & 0.843 \\
11 & 0.95 & 0.05 & 0.5 & $0.9522\pm0.0049$ & 5.13 & 1.213 & 0.843 \\
12 & 0.8  & 0.05 & 0.5 & $0.9482\pm0.0023$ & 5.22 & 1.235 & 0.830 \\
13 & 0.9  & 0.1  & 0.3 & $0.9474\pm0.0060$ & 5.26 & 1.244 & 0.822 \\
14 & 0.95 & 0.1  & 0.7 & $0.9526\pm0.0001$ & 5.42 & 1.281 & 0.812 \\
15 & 0.9  & 0.05 & 0.7 & $0.9474\pm0.0060$ & 5.40 & 1.277 & 0.806 \\
16 & 0.8  & 0.05 & 0.7 & $0.9482\pm0.0023$ & 5.47 & 1.295 & 0.800 \\
17 & 0.95 & 0.02 & 0.7 & $0.9509\pm0.0030$ & 5.52 & 1.307 & 0.796 \\
18 & 0.95 & 0.05 & 0.7 & $0.9522\pm0.0049$ & 5.64 & 1.335 & 0.782 \\
19 & 0.8  & 0.02 & 0.7 & $0.9482\pm0.0023$ & 5.63 & 1.332 & 0.781 \\
20 & 0.8  & 0.1  & 0.5 & $0.9482\pm0.0023$ & 5.80 & 1.371 & 0.761 \\
21 & 0.95 & 0.02 & 0.3 & $0.9528\pm0.0009$ & 5.93 & 1.402 & 0.751 \\
22 & 0.95 & 0.02 & 0.5 & $0.9509\pm0.0030$ & 5.94 & 1.405 & 0.747 \\
23 & 0.95 & 0.1  & 0.3 & $0.9526\pm0.0001$ & 6.05 & 1.431 & 0.737 \\
24 & 0.8  & 0.02 & 0.3 & $0.9482\pm0.0023$ & 6.01 & 1.421 & 0.737 \\
25 & 0.9  & 0.1  & 0.7 & $0.9474\pm0.0060$ & 6.05 & 1.431 & 0.729 \\
26 & 0.8  & 0.1  & 0.3 & $0.9482\pm0.0023$ & 6.24 & 1.477 & 0.709 \\
27 & 0.95 & 0.05 & 0.3 & $0.9522\pm0.0049$ & 6.44 & 1.525 & 0.687 \\
\bottomrule
\end{tabular}
\end{table*}
To evaluate the robustness of our self-enhancement classifier under different hyperparameter settings, we conducted a full-factor grid search over the three key parameters: initial threshold \(\tau_0 \in \{0.8, 0.9, 0.95\}\), decay rate \(\beta \in \{0.02, 0.05, 0.1\}\), and lower bound \(\tau_{\min} \in \{0.3, 0.5, 0.7\}\). This resulted in 27 configurations. For each configuration, we measured the macro F1 score (with standard deviation) and the average inference latency on the same validation set. To comprehensively evaluate each hyperparameter configuration, we define a combined score that balances accuracy, stability, and efficiency. Let \(\mu_{\text{lat}}(c)\) denote the average inference latency of configuration \(c\), and let \(\mu_{\text{lat}}^{\min}\) be the minimum average latency among all candidate configurations:
\begin{equation}
\mu_{\text{lat}}^{\min} = \min_{c} \mu_{\text{lat}}(c).
\end{equation}
The normalized latency is then defined as:
\begin{equation}
\mathrm{LatencyNorm}(c) = \frac{\mu_{\text{lat}}(c)}{\mu_{\text{lat}}^{\min}}.
\end{equation}
With penalty coefficients \(\lambda_s = 0.5\) and \(\lambda_l = 0.5\), the overall score for configuration \(c\) is:
\begin{equation}
\begin{split}
\mathrm{Score}(c) = &\; \mu_{\mathrm{F1}}(c) - \lambda_s \cdot \sigma_{\mathrm{F1}}(c)- \\
& \lambda_l \cdot \max\!\bigl(0,\; \mathrm{LatencyNorm}(c)\bigr)
\end{split}
\end{equation}
where \(\mu_{\text{F1}}(c)\) and \(\sigma_{\text{F1}}(c)\) are the mean and standard deviation of the macro F1 score, respectively. This score penalizes both performance variability (\(\sigma_{\text{F1}}\)) and any latency exceeding the optimal value (\(\mathrm{LatencyNorm}(c)-1\)), thereby encouraging configurations that are accurate, stable, and efficient.

As shown in the table~\ref{tab:sensitivity}, the macro F1 values across all 27 configurations vary within a very small range, demonstrating that the classifier is remarkably insensitive to the choice of these hyperparameters. The best macro F1 (0.9528) is achieved by configuration \#21 (\(\tau_0=0.95, \beta=0.02, \tau_{\min}=0.3\)), but its latency (5.93~ms) is significantly higher than the optimum. The configuration with the lowest latency (4.23~ms) is \#1 (\(\tau_0=0.9, \beta=0.02, \tau_{\min}=0.5\)), which also attains a macro F1 of 0.9490.

Our default configuration \#5 (\(\tau_0=0.9, \beta=0.05, \tau_{\min}=0.5\)) yields a macro F1 of 0.9474 and a latency of 4.44~ms, which is only marginally worse than the best trade-off. The differences from the top-scoring configuration (\#1) are minimal: macro-F1 differs by only 0.0016 (0.9474 vs. 0.9490) and latency by 0.21~ms (4.44 vs.4.23). Both configurations lie on the same near-Pareto frontier. We choose \#5 over \#1 for the following reasons:
\begin{itemize}[leftmargin=*]
\item \textbf{Adaptive threshold updates}: The moderate decay rate (\(\beta=0.05\)) triggers more effective threshold updates during online self-enhancement, whereas a smaller decay (\(\beta=0.02\)) would keep the threshold nearly unchanged, leading to under-adaptation.
\item \textbf{Consistency with our experimental protocol}: This setting avoids post-hoc overfitting to a single best point and maintains consistency with the rest of our experiments.
\item \textbf{Robustness}: The performance gap between \#5 and the best configuration is negligible, and our main conclusions remain unchanged regardless of which near-optimal configuration is chosen.

Therefore, we adopt \(\tau_0=0.9, \beta=0.05, \tau_{\min}=0.5\) as the default setting for all main experiments. The full ranking of 27 configurations is provided to demonstrate that our results are insensitive to the specific choice among near-optimal settings.
\end{itemize}

\renewcommand{\floatpagefraction}{.9}
\begin{figure*}[t]
    \centering
    \includegraphics[width=\textwidth]{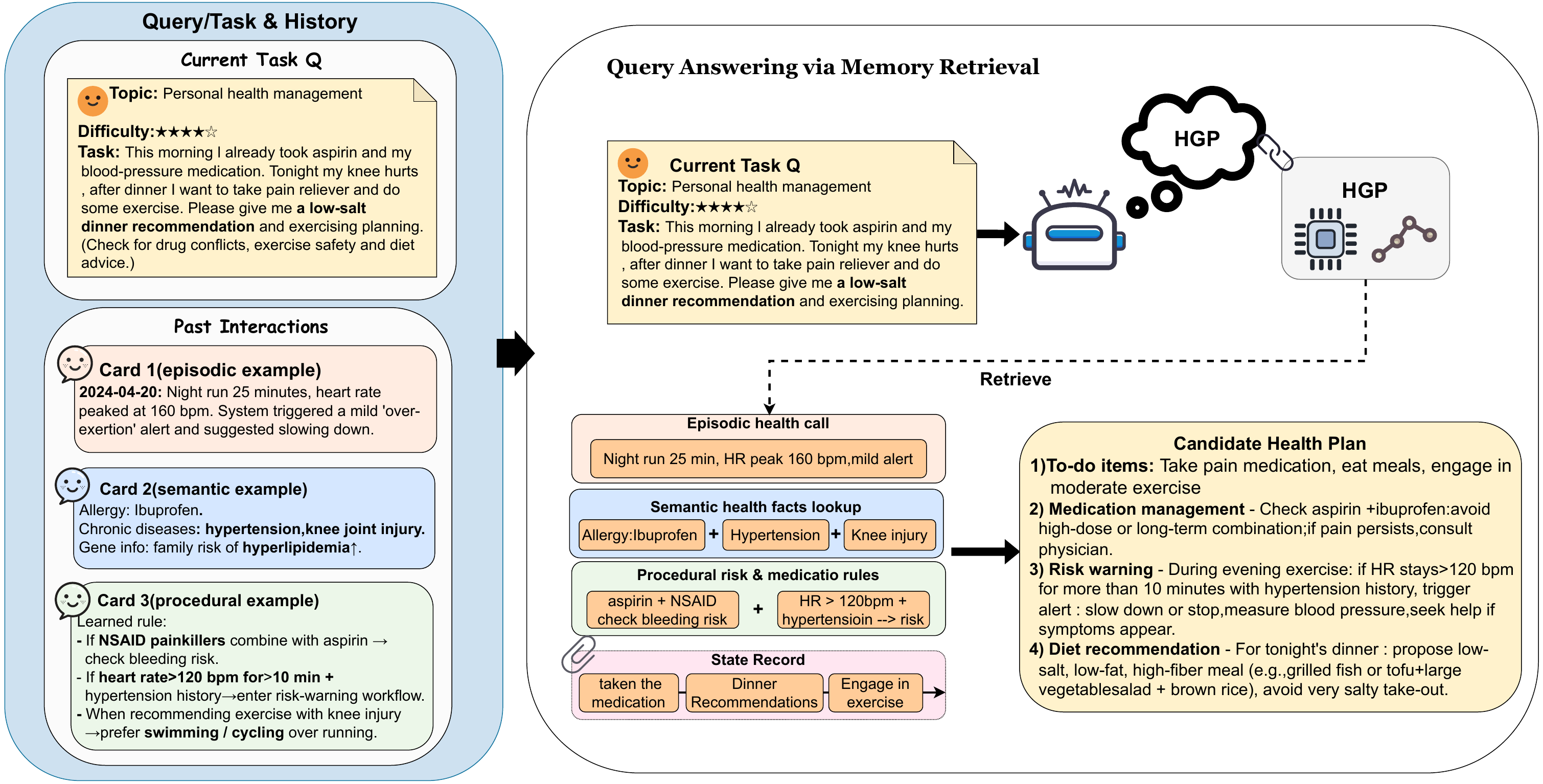}
    \vspace{-4mm}
    \caption{An illustrative example of \model in a personalized health management scenario. Given a user’s current query and long-term interaction history, HGP organizes heterogeneous memories, including episodic, semantic, procedural, and working memory, within a hybrid graph storage, and retrieves task-relevant evidence to generate personalized health recommendations.
}
    \label{fig:hgp_case}
    \vspace{-4mm}
\end{figure*}
\subsubsection{Stage-wise Performance of the Self-Enhancement Classifier}
\begin{table}[htbp]
\centering
\small
\setlength{\tabcolsep}{4pt}   
\caption{Performance evolution of the self-enhancement classifier across incremental training stages.}
\label{tab:stage_results}
\begin{tabular}{lccc}
\toprule
\textbf{Stage} & \textbf{Macro-F1} & \textbf{Micro-F1} & \textbf{\(\Delta\)Macro (vs. S0)} \\
\midrule
S0 (offline init) & 0.8980 & 0.8859 & 0.0000 \\
S1 & 0.9162 & 0.9068 & +0.0182 \\
S2 & 0.9282 & 0.9192 & +0.0302 \\
S3 & 0.9445 & 0.9275 & +0.0465 \\
S4 & 0.9527 & 0.9379 & +0.0547 \\
S5 & 0.9554 & 0.9416 & +0.0574 \\
\bottomrule
\end{tabular}
\end{table}
To investigate how the self-enhancement mechanism improves classifier performance over time, we record its performance after each incremental training round. The initial stage (S0) uses an offline-trained classifier without any LLM supervision. Subsequently, every 200 newly LLM-labeled samples trigger one round of incremental fine-tuning, resulting in five rounds (S1 to S5). As shown in Table~\ref{tab:stage_results}, the macro F1 steadily increases from 0.8980 (S0) to 0.9554 (S5), an absolute gain of 5.74 percentage points. The micro F1 follows a similar upward trend. The most significant improvements occur in the first two rounds (S0$\rightarrow$S1: +1.82 points; S1$\rightarrow$ S2: +1.20 points), after which the gains gradually diminish, indicating convergence. These results demonstrate that the self-enhancement mechanism effectively leverages LLM-labeled hard samples to continuously improve routing accuracy without causing performance oscillation or catastrophic forgetting.

It is important to emphasize the training protocol for this stage-wise evaluation. The classifier first makes predictions on the training set. Samples with low confidence are sent to an LLM for re-labeling, and the (sample, label) pairs are cached. Once a sufficient number of such samples is accumulated, incremental training is performed, and the cycle repeats. The training and test sets are strictly separated. The test set is never used for training, nor does it ever involve LLM inference. For rigorous evaluation, the test set is further partitioned into multiple disjoint subsets, each used exactly once for evaluating the classifier after one incremental training round. This prevents any data leakage across evaluation steps.

\section{Retrieval Workflow Example}
\label{appendix:retrieval_example}
As shown in Figure~\ref{fig:hgp_case}, upon receiving a user query \( q \) , \model\ executes a structured retrieval protocol over its hybrid graph memory \( \mathcal{G} = \{ \mathcal{G}_{\mathrm{epi}}, \mathcal{G}_{\mathrm{sem}}, \mathcal{G}_{\mathrm{proc}}, \mathcal{G}_{\mathrm{wm}} \} \). Each sub-graph is queried for a specific type of evidence. The retrieved sub-graphs are merged into a unified context and passed, together with \( q \), to the LLM. The model then generates a response that is aware of both long-term user history and the current implicit constraints, thereby ensuring personalization, safety, and coherence.

\section{Details of special case}
\begin{figure*}[t]
    \centering
    \includegraphics[width=\textwidth]{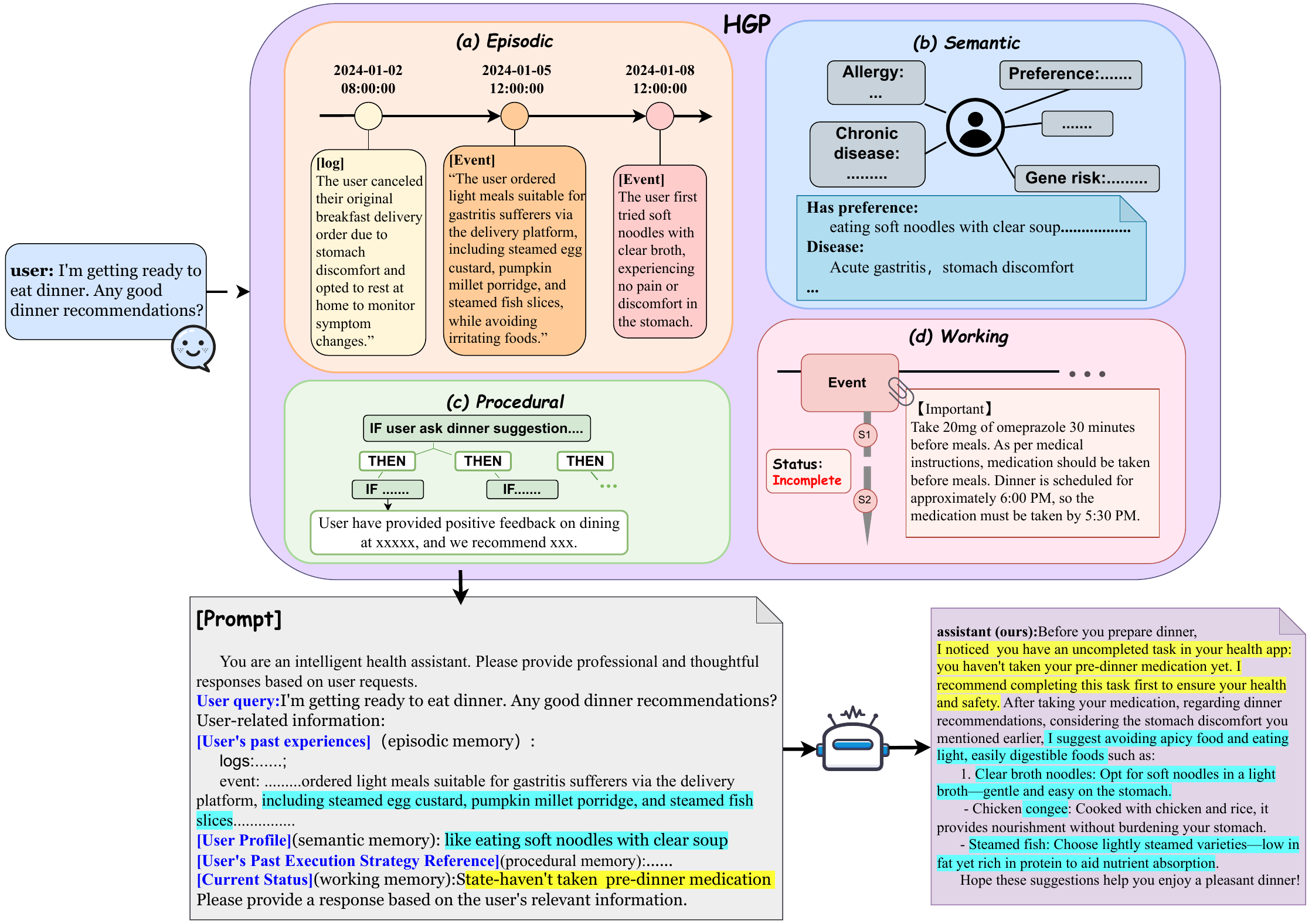}
    \vspace{-4mm}
    \caption{Special case study: illustrating the detailed process by which an LLM retrieves user-relevant memories via HGP in the special case. Working memory addresses \emph{implicit constraints} in a personalized health assistant scenario; its effect is highlighted in yellow (pre-meal medication: pending), while personalized recommendation cues are highlighted in blue.
}
    \label{fig:case_detail}
    \vspace{-4mm}
\end{figure*}
\label{appendix:special_case}
We construct a personalized health scenario where the dialogue history includes a to-do item: \emph{take pre-meal medication before dinner}. During evaluation, the user only requests dinner recommendations without explicitly mentioning medication. This case tests whether the agent can adhere to \emph{implicit constraints} inferred from prior interactions and prioritize actions that are required but not explicitly stated in the query.

Figure~\ref{fig:case_detail} illustrates the full pipeline for the case study in Section~\ref{sec:special_case_study}. During retrieval, \model queries heterogeneous external memories separately: (i) the event graph for diet- and symptom-related logs (episodic memory); (ii) the knowledge graph for user preferences and profile facts (semantic memory); and (iii) the rule tree for reusable strategies that previously received positive feedback (procedural memory). Together, these long-term memories supply user-relevant evidence for personalized dinner suggestions (highlighted in blue).

Crucially, \model also queries the \emph{working-memory states} from the state trajectory graph to retrieve the current actionable status.
In this case, working memory returns ``pre-meal medication: incomplete'' and injects it as an \emph{actionable constraint} into the prompt. Then the LLM reminds the user to take the medication before offeringrecommendations (highlighted in yellow).This behavior is difficult to trigger reliably using only episodic, semantic, or procedural retrieval, as it hinges on short-horizon execution state rather than long-term facts, preferences, or strategies.

Overall, this case demonstrates \model’s ability to deliver personalized, context-aware responses. It also shows that long-term memory retrieval alone cannot guarantee compliance with implicit, actionable constraints. By explicitly tracking short-term states via working memory, \model enhances behavioral consistency and safety in long-horizon interactions.

\section{Abstraction Operators for Semantic and Procedural Memory from Episodic Memory (LLM Prompts)}
To construct semantic and procedural memory from raw episodic traces, we design two abstraction operators, \(\mathcal{A}_{\mathrm{sem}}\) and \(\mathcal{A}_{\mathrm{proc}}\), each implemented via a dedicated LLM prompt template. These operators extract structured knowledge from episodic event graphs and convert them into reusable memory formats.
\subsection{Semantic Abstraction Operator \(\mathcal{A}_{\mathrm{sem}}\)}
\label{appendix:prompt_sem}
\begin{figure*}[t]
    \centering
    \setlength{\abovecaptionskip}{-4mm}
    \includegraphics[width=\textwidth]{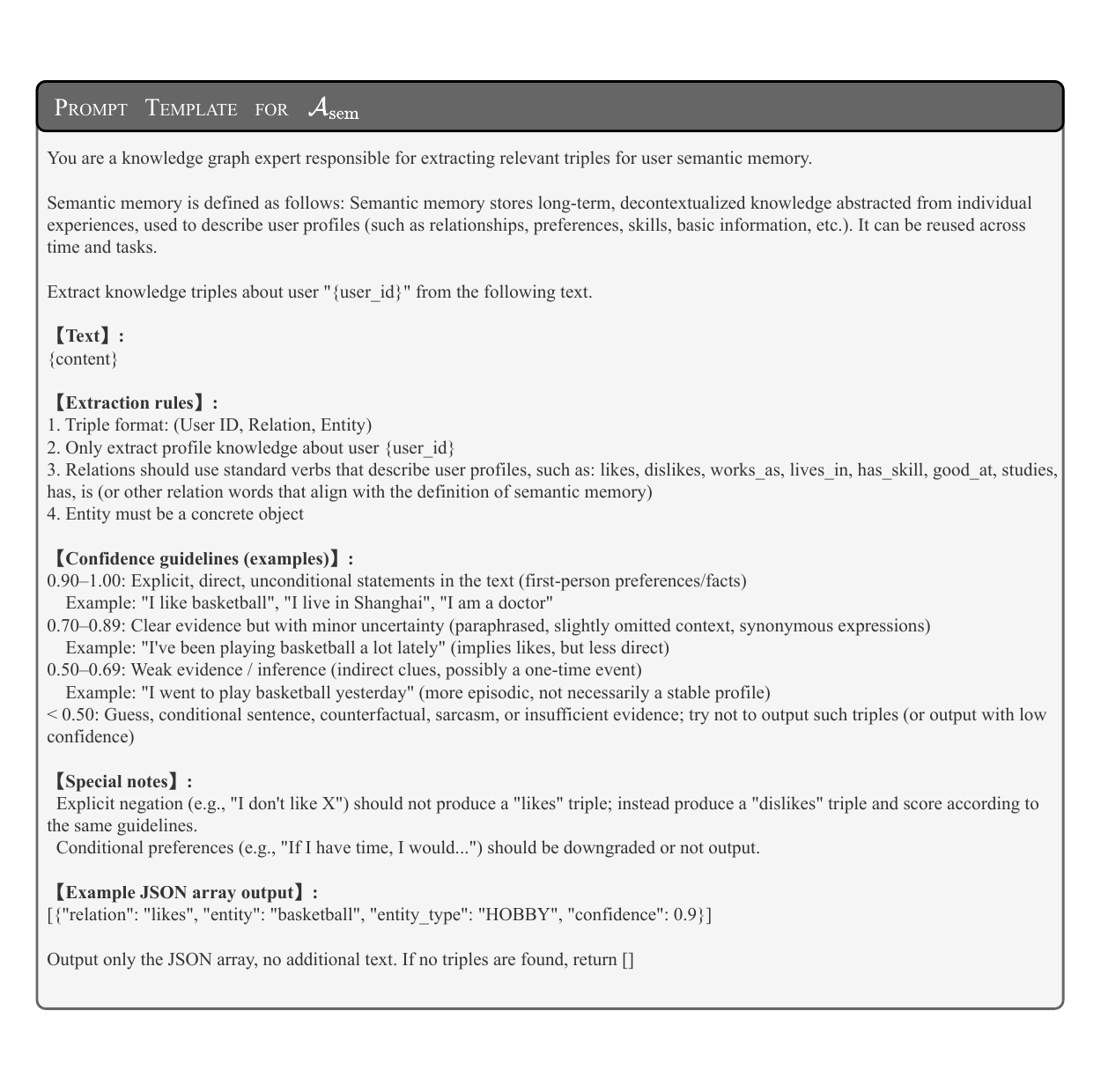}
    \caption{Illustration of the procedural abstraction operator $\mathcal{A}_{\mathrm{sem}}$.}
    \label{fig:prompt_sem}
\end{figure*}

The semantic abstraction operator \(\mathcal{A}_{\mathrm{sem}}\) extracts static factual knowledge from episodic interactions to construct user profiles, including user attributes, preferences, skills, and long-term constraints. As shown in Figure~\ref{fig:prompt_sem}, \(\mathcal{A}_{\mathrm{sem}}\) employs a structured LLM prompt to identify and summarize semantic triples in the form of a knowledge graph, given episodic memories.
\subsection{Procedural Abstraction Operator \(\mathcal{A}_{\mathrm{proc}}\)}
\label{appendix:prompt_proc}
\begin{figure*}[htbp]
    \centering
    \setlength{\abovecaptionskip}{-4mm}
    \includegraphics[width=\textwidth]{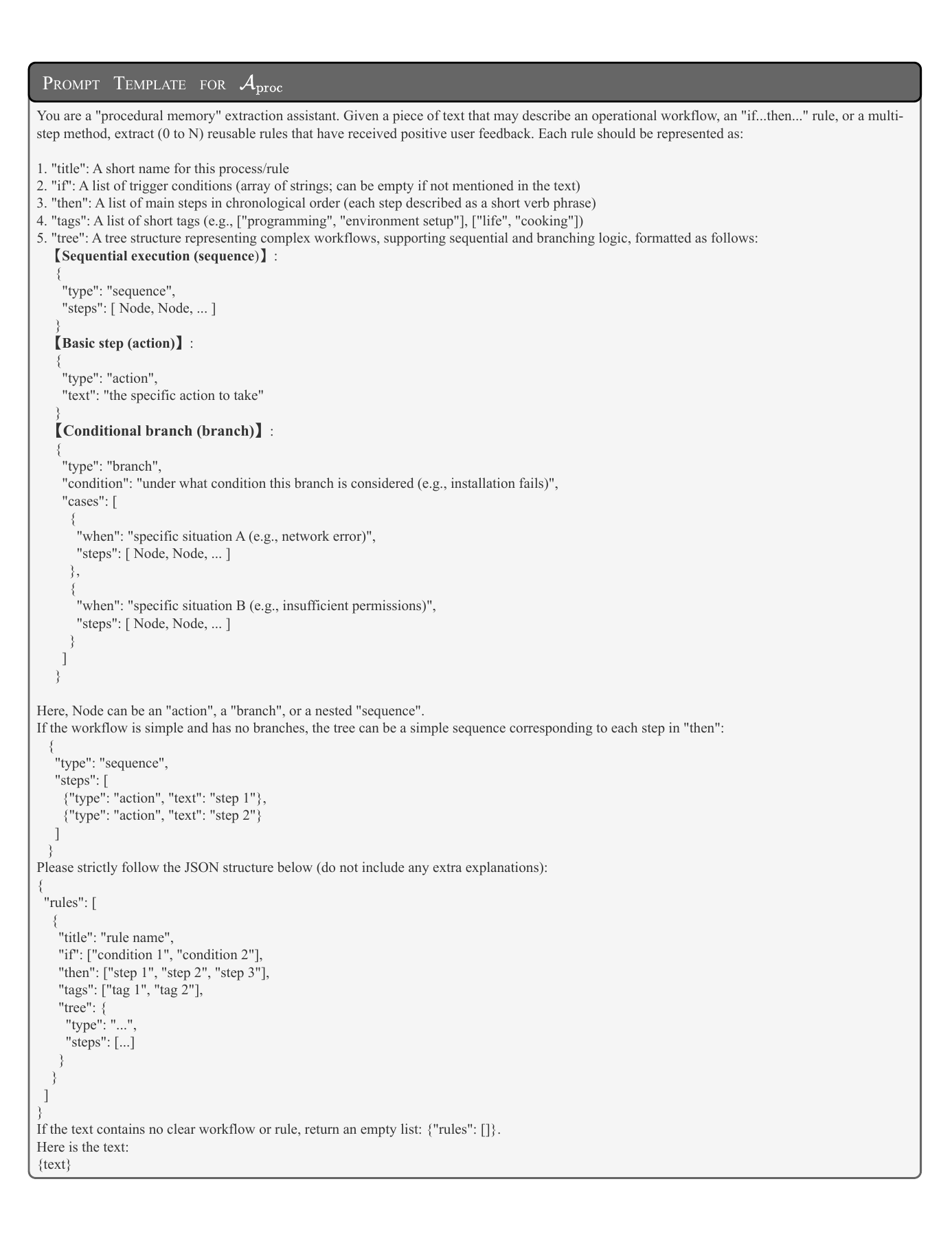}
    \caption{Illustration of the procedural abstraction operator $\mathcal{A}_{\mathrm{proc}}$.}
    \label{fig:prompt_proc}
\end{figure*}

The procedural abstraction operator \(\mathcal{A}_{\mathrm{proc}}\) extracts reusable rules, workflows, and decision strategies from episodic traces. As illustrated in Figure~\ref{fig:prompt_proc}, \(\mathcal{A}_{\mathrm{proc}}\) uses a structured LLM prompt that requires the model to output each rule in a unified JSON format, containing fields such as title, trigger conditions, execution steps, tags, and a tree-structured workflow that supports sequential and branching logic. If no clear workflow is present, the operator returns an empty list. The extracted rules are then stored in the procedural memory graph \(\mathcal{G}_{\mathrm{proc}}\).

\end{document}